\def\year{2020}\relax
\documentclass[letterpaper]{article} 
\usepackage{aaai20}  
\usepackage{times}  
\usepackage{helvet} 
\usepackage{courier}  
\usepackage[hyphens]{url}  
\usepackage{graphicx} 
\usepackage{graphicx}  
\graphicspath{ {AuthorKit20/LaTeX/images/} }
\usepackage{amsmath}
\usepackage{amsthm}
\usepackage{algorithm}
\usepackage{algpseudocode}
\usepackage{subfig}
\newcommand{\squeezeup}{\vspace{-2.5mm}}
\title{Using Explainable Scheduling for the Mars 2020 Rover Mission}
\author{Jagriti Agrawal, Amruta Yelamanchili, Steve Chien\\ 
Jet Propulsion Laboratory\\
California Institute of Technology\\
4800 Oak Grove Drive\\
Pasadena, California 91109\\
firstname.lastname@jpl.nasa.gov 
}
 
\begin{document}

\maketitle

\begin{abstract}

Understanding the reasoning behind the behavior of an automated scheduling system is essential to ensure that it will be trusted and consequently used to its full capabilities in critical applications. In cases where a scheduler schedules activities in an invalid location, it is usually easy for the user to infer the missing constraint by inspecting the schedule with the invalid activity to determine the missing constraint.  If a scheduler fails to schedule activities because constraints could not be satisfied, determining the cause can be more challenging. In such cases it is important to understand which constraints caused the activities to fail to be scheduled and how to alter constraints to achieve the desired schedule. In this paper, we describe such a scheduling system for NASA's Mars 2020 Perseverance Rover, as well as Crosscheck, an explainable scheduling tool that explains the scheduler behavior. The scheduling system and Crosscheck are the baseline for operational use to schedule activities for the Mars 2020 rover. As we describe, the scheduler generates a schedule given a set of activities and their constraints and Crosscheck: (1) provides a visual representation of the generated schedule;  (2) analyzes and explains why activities failed to schedule given the constraints provided; and (3) provides guidance on potential constraint relaxations to enable the activities to schedule in future scheduler runs. 
\end{abstract}

\section{Introduction}

In order for an automated scheduling system to be used for highly critical applications, such as space missions, the users must both trust the system and understand its behavior. One such application is the Mars 2020 (M2020) Perseverance Rover, which launched July 30, 2020 and is expected to land on Mars of February 18, 2021 \cite{mars2020}. The mission will use a ground-based scheduler, called Copilot, with plans to transition to using an onboard scheduler later on in the mission. After the transition, the ground scheduling tools will be used to predict and analyze how the onboard scheduler may behave with a given input plan, as well aid in the reconstruction of onboard scheduling/execution traces. Both the ground scheduler and the onboard scheduler  use the same core scheduling algorithm \cite{rabideau-benowitz-IWPSS-2017} \cite{chi-icaps2018-embedding} \cite{agrawal-iwpss2019-disjunction}
\cite{chi-icaps2020-wakesleep}.

Such automated scheduling systems have not been used for prior NASA planetary rover missions (with the notable exception of the Mars Exploration Rover Mission \cite{bresina2005activity}). Instead, science team members created the activities for the rover and science planners had to manually generate a schedule consisting of the activities and ensure that the schedule satisfied all given constraints.  

This process resulted in a significant amount of time being spent to create the schedule. In addition, the activity durations provided were very conservative. Data from the Mars Science Laboratory Mission \cite{gaines2016productivity,gaines2016tech} indicates that activities completed with durations on average 28\% shorter than predicted. The ground scheduler primarily aims to significantly reduce the time spent to generate a correct schedule. Since the onboard planner will be able to account for activities taking more or less time than expected \cite{chi-icaps2018-embedding}, the onboard planner will allow the science team to be less conservative when determining activity durations, resulting in higher mission productivity.

\par
Using a scheduler that can generate a schedule which will satisfy the constraints provided by the science team will save much needed time in the process of creating a schedule for the rover, as well as increase mission productivity. In order to build a high level of confidence and trust in such a system, it is essential to have a tool that can easily explain why the final schedule looks the way it does, particularly when a critical activity was not able to be scheduled due to the provided constraints. 

Crosscheck is an explainable scheduling tool that provides a visual representation of the generated schedule and explains exactly why certain activities failed to schedule. By using Crosscheck, users will not only gain confidence in the scheduling system but they will also understand how they can alter constraints to achieve a desired schedule.

The remainder of the paper is organized as follows. First, we describe the scheduling problem and what types of constraints the scheduler must satisfy. We also discuss how the scheduler auto-generates and schedules certain activities that are required by the activities given by the science planners. These auto-generated activities ensure the rover is sufficiently heated to perform activities, determines when the CPU must be on to perform activities, and when the CPU can be turned off to preserve energy. Second, we describe the algorithms used in Crosscheck to identify why a given activity failed to be schedule and how the constraints may be altered to achieve the desired schedule. Third we provide visual examples of what kinds of failure causes Crosscheck is able to identify. We also describe user feedback on Crosscheck. Finally, we discuss related work in this field as well as areas for future work.

\section{Problem Definition and Scheduling Constraints}

The target scheduler is a non-backtracking scheduler to be onboard the M2020 rover \cite{rabideau-benowitz-IWPSS-2017} that schedules in priority first order and never removes or moves an activity after it is placed during a single run of the scheduler. The priorities of each activity are initially set using heuristics and then adjusted using a squeaky wheel approach described in \cite{chi-icaps2019-optimizing}. 
For the scheduling problem we adopt the definitions in  \cite{rabideau-benowitz-IWPSS-2017}. The scheduler is given a list $A$ of activities with their constraints,  $A_{1}\langle p_{1}, d_{1}, R_{1}, e_{1}, dv_{1}, \Gamma_{1}, T_{1}, D_{1} \rangle \ldots \\
A_{n} \langle p_{n}, d_{n}, R_{n}, e_{n}, dv_{n}, \Gamma_{n} ,T_{n}, D_{n} \rangle$  where 

\begin{itemize}
\item $p_{i}$ is the scheduling priority of activity $A_i$. A higher numerical value for $p_{i}$ indicates a lower scheduling priority and vice versa.
\item $d_{i}$ is the nominal, or predicted, duration of activity $A_i$.
\item $R_{i}$ is the set of unit resources $R_{i_1} \ldots R_{i_m}$ that activity $A_i$ will use. Two activities cannot be scheduled in parallel if they use the same unit resource. For example, if two activities both claim the Arm resource, then they cannot be scheduled in parallel.
\item $e_{i}$ and $dv_{i}$ are the average rates at which the consumable resources energy and data volume respectively are consumed by activity $A_i$.
\item $peak_{i}$ is the maximum rate at which the activity will consume energy at any point.
\item $\Gamma_{i_1} \ldots \Gamma_{i_r}$ are non-depletable resources used such as sequence engines available or peak power for activity $A_i$.
\item $T_i$ is a set of allowed start time windows 
$\lbrack T_{i_{j\_start}}$,  $T_{i_{j\_preferred}}$, $T_{i_{j\_end}}$ \rbrack \ldots \lbrack$T_{i_{k\_start}}$, $T_{i_{k\_preferred}}$, $T_{i_{k\_end}}\rbrack$ 
for activity 
$A_i$. Activity ${A_i}$ must be scheduled within one of its allowed start time windows. \footnote{If a preferred start time, $T_{i_{j\_preferred}}$ is not specified for window $j$ then it is by default $T_{i_{j\_start}}$}.
\item $D_i$ is a set of activity dependency constraints for activity $A_i$ where $A_p \rightarrow A_q$ means $A_q$ must execute successfully before $A_p$ starts.
\item $SR_i$ is a set of activity state constraints or requirements for activity $A_i$ where a certain rover state must be satisfied before activity $A_i$ can execute. The rover state may be satisfied either by a previous activity that sets the state to the necessary value or it may be satisfied if the incoming rover state is the required value.

\item $SE_i$ is a set of activity state effects for activity $A_i$ where a certain rover state is set to a value by activity $A_i$. This rover state may be required by another activity in order for it to execute.

\item $U_i$ is a set of UHF interactions for activity $A_i$. A UHF interaction determines which kinds of specific data uplink and downlink activities $A_i$ are allowed to be scheduled in parallel with.

\end{itemize}

Each activity may also require the automatic generation of:
1) a set of preheat activities, $P_i = \{p_{i_1} \ldots p_{i_k}\}$, 
2) a set of maintenance heating activities, $M_i = \{m_{i_1} \dots m_{i_k}\}$, 
or 3) an awake activity, $a_i$.
Preheats are setup activities (i.e. they occur before the activity), 
while maintenance heating and awakes are companion activities (i.e. they occur during or with the activity).

In addition to satisfying the constraints listed above for each individual activity, the scheduler must also ensure that scheduling an activity will not violate certain global constraints. These global constraints include the following.
\begin{itemize}
    \item \textit{Minimum State of Charge (SOC) ($C\ _{soc}^{min}$)} The state of charge, or energy value, cannot dip below the Minimum SOC at any point. If scheduling an activity would cause the energy value to dip below the Minimum SOC, then that activity will not be scheduled.

     \item \textit{Maximum Peak Power ($C\ _{p\_power}^{max}$)}  If scheduling the activity would result in the total peak power at a given time point to exceed the Maximum Peak Power, then the activity will not be scheduled.

\end{itemize}

\section{Auto-generated Activities}

\begin{figure}[h!]
    \centering
    \includegraphics[width=0.75\linewidth, keepaspectratio]{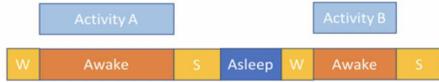}
    \caption{The scheduler must schedule periods when the rover will be awake and asleep while satisfying the minimum awake/asleep requirements.}
    \label{fig:sleep-scheduling}
\end{figure}

The M2020 rover’s power source constantly generates energy. 
However, the CPU’s awake and “idle” state (i.e. no other activities) consumes more energy than the source provides. Therefore, the rover's energy, or battery SOC, only increases when it is asleep. The rover, however, must be awake to execute certain activities. The scheduler is responsible for generating and scheduling these periods when the rover must wakeup, remain awake, shut down, and then sleep as shown in Figure \ref{fig:sleep-scheduling}. While scheduling these awake and asleep periods, the scheduler must consider the following types of constraints to prevent the rover from waking up and shutting down too frequently.
\begin{itemize}
\item \textit{Minimum Sleep Time($C\ _{sleep}^{min}$)} Any given asleep period must be at least a certain duration. 
\item \textit{Minimum Awake Time ($C\ _{awake}^{min}$)} Any given awake period must be at least a certain duration. 
\end{itemize}

More details about how these awake and asleep periods are generated can be found in \cite{chi-icaps2020-wakesleep}.

\begin{figure}[h!]
    \centering
    \includegraphics[width=0.85\linewidth, keepaspectratio]{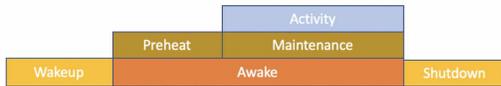}
    \caption{The scheduler must schedule the required heating for a given activity as well.}
    \label{fig:heat-scheduling}
\end{figure}

Certain activities may require an instrument on the rover to be sufficiently heated before the activity can execute and this heating must be maintained for the duration of the activity. For a given activity, $A_i$, the scheduler must autogenerate and schedule these heating activities called preheats, $P_i$ which must complete prior to the start time of an activity, and maintenances, $M_i$, which occur during the activity, as shown in Figure \ref{fig:heat-scheduling}. If the required preheats and maintenances are not able to be scheduled, then the activity will also not be able to be scheduled.

These autogenerated preheats and maintenances consume energy just as other activities do. The duration of the preheat is dependent on the time of day. During times when the temperature is colder, a longer preheat will be required than during times when the temperature is warmer. If the preheat is too long and consumes too much energy then it is possible it will not be able to scheduled due to violation of $C\ _{soc}^{min}$. Preheats also have defined \textit{operability windows}. If the activity is scheduled such that its preheat would need to start outside of its operability window, then the preheat and activity will not be able to scheduled. 

If any of the above constraints are violated while the scheduler is attempting to schedule the activity or any of its required autogenerated awake, asleep, or heating activities then the activity will not be scheduled. Since all input activities coming from the user are deemed \textit{mandatory}, if science planners see that an activity failed to schedule they will want to know why. 

In the following sections, we discuss how Crosscheck analyzes the scheduling behavior described above to identify why an activity failed to schedule and how it presents this information to users in a comprehensible way.

\section{Determining First Scheduling Step when Constraints Failed to be Satisfied}

After the scheduler generates a schedule with the given activities and constraints, users will use Crosscheck to view and understand the resulting schedule. Crosscheck analyzes the behavior of the scheduler to provide additional information while each activity is being scheduled in order to identify which constraints caused the activity to fail to schedule.  

The onboard scheduler that will be used further into the mission has extremely limited CPU resources available\footnote{The RAD750 processor used by the Mars 2020 rover has measured performance in the 200-300 MIPS range.  In comparison a 2016 Intel Core i7 measured over 300,000 MIPS or over 1000 times faster.  Furthermore, the onboard scheduler is only allocated a portion of the computing cycles onboard the RAD750 resulting computation {\em several thousand times} slower than a typical laptop.} and a conservative estimated runtime of 60 seconds. Backtracking would result in a significantly higher runtime. Thus it does not backtrack. Since it is crucial that the ground scheduler, which will be used immediately after landing, sufficiently prepares the team for use of the onboard scheduler it is important that both schedulers use the same scheduling algorithm. Therefore, the ground scheduler also does not backtrack. 

Since the scheduler does not backtrack while scheduling an activity, the order in which activities are scheduled, given by the scheduling priority, can have a significant impact on the final schedule.

Each time the scheduler attempts to schedule an activity, it performs a new scheduling step. Consequently, the total number of scheduling steps is equal to the number of activities that must be scheduled. We define a \textit{failure step} of a given activity as a step during which the activity is scheduled which results in the activity failing to be scheduled. The goal is to determine the earliest failure step of an activity that failed to schedule given the initial input plan.

Changing constraints of activities with a scheduling priority that would result in them being scheduled after the first failure step will not change whether the activity is able to schedule. However, it is possible that once constraints change for activities scheduled at or before the first failure step that is identified, activities scheduled after the first failure step will contribute to the failure or success of the activity in question, and a subsequent viewing of the updated schedule in Crosscheck would reveal these. 

Crosscheck finds the earliest failure step of a given activity by using a binary search method. Algorithm \ref{alg:binary-search} describes how Crosscheck finds the earliest failure step for an activity and the resulting partial schedule.

{
\setlength\intextsep{2pt}
\begin{algorithm}[t]
\small
  \caption{Binary Search to Find Failure Step}
  \label{alg:binary-search}
  \begin{algorithmic}[1]
    \Require
    \Statex $A$: List of activities with their individual constraints
    \Statex $P$: Plan wide constraints (e.g. available cumulative
    resources)
    \Statex $p_f$: priority of activity that failed to schedule
    \Ensure
    \Statex $Partial_{sc}$: Partial schedule consisting of activities scheduled in the failure step
    \State $low \gets 1$
    \State $high \gets len(A)$
    \While {$low \neq high$}
      \State $Partial_{sc} \gets \{\}$
      \State $Partial_{in} \gets \{\}$
      \State $p_{curr} \gets low + (high - low \div 2)$
      \ForAll {$A_i \in A $}
         \If {$p_i \leq p_{curr}$}
            \State $Partial_{in}.insert(A_i)$
         \EndIf
      \EndFor
      $Partial_{sc} \gets run\_scheduler(Partial_{in})$
      
      \If {$A_i \notin Partial_{sc}$}
         \State $high \gets p_{curr}$
      \Else
         \State $low \gets p_{curr} + 1$
      \EndIf
      
    \EndWhile

  \end{algorithmic}
\end{algorithm}
}

If an activity $A_f$ with scheduling priority $p_f$  is scheduled at step $p_f$ in the initial plan, then it is first considered for scheduling by Crosscheck at step ${p_f}/2$ = $p_f'$. All activities with priorities 0 through $p_f'-1$ are included in the plan. If the activity $A_f$ successfully schedules in this first partial input plan, then the first failure step is due to an activity that with a lower scheduling priority (higher numerical value for $p_i$). The process repeats, with the steps under consideration being between $p_f'$ and $p_f$. If the activity $A_f$ is not successfully scheduled, the process is repeated with the steps under consideration being between 1 and $p_f'$. This process is repeated until the earliest failure step is found.

Some constraints that $A_f$ has may be dependent on other activities being present in the plan. These include dependency constraints and state requirements. A constraint that falls in either of these two categories is only included in the current partial input plan if all activities that could satisfy the constraint and having a higher scheduling priority are also present in the input plan. 

Dependency constraints are tied to a single other activity. If that activity is in the plan, then the dependency constraint is in the plan. If an activity $A_{f}$ has a dependency on activity $A_{i}$, then $A_{i}$ will have a higher scheduling priority than $A_f$.  

State requirements may be satisfied by the effect of one of many activities. A state requirement constraint is not included in a plan unless all the activities with a higher scheduling priority than that of a $A_f$ that have a state effect that could satisfy the requirement are included in the plan. Consider the following example. Activity $A_3$ with scheduling priority $p_3$ may have a state requirement $SR_3$ that could be satisfied by either state effect $SE_1$ given by activity $A_1$ with priority $p_1$, or by $SE_2$ given by activity $A_2$ with priority $p_2$. Activities $A_1$ and $A_2$ have a higher scheduling priority than activity $A_3$. If requirement $SR_3$ is added to the partial input plan consisting of the activities $A_1$ and $A_3$ then $A_3$ may fail to schedule if the activity $A_1$ fails to schedule. It would be premature to determine step 1 as the earliest failure step, as the activity may be able to successfully schedule after $A_2$ is scheduled. Thus, the requirement is not be added to the plan unless both the activities $A_1$ and $A_2$ are present in the plan. 

\section{Determining which Constraints Failed to be Satisfied}

Once we find the failure step of an activity that failed to schedule, we can analyze why it failed to schedule.
The scheduling algorithm consists of two primary phases, and an activity can fail to schedule for various reasons in either of these phases. The two phases are described as follows. 

\begin{enumerate}
    \item We find \textit{valid intervals} for each activity constraint. Similar to the valid intervals used in scheduling for the Rosetta Orbiter \cite{chien-rabideau-tran-et-al-IJCAI-2015}, we define a \textit{valid interval} for a given constraint as a continuous time interval within the plan bounds where the constraint will be satisfied for the duration of the activity. The valid intervals for each constraint are intersected with each other to generate the activity's \textit{final valid intervals}. If the set of final valid intervals for an activity is empty, then the activity fails to schedule. Valid constraint intervals were also calculated while scheduling for the Rosetta Orbiter.
    \item We account for plan wide energy constraints, wake sleep activities, and heating activities in order to determine the activity's final start time. If the algorithm fails to find a time within the final valid intervals that satisfies these additional requirements, then the activity fails to schedule. 
    
\end{enumerate}

Crosscheck determines the cause of an activity failing to schedule in either of the two phases of the scheduling algorithm.

\subsection{Failure due to Constraint Valid Intervals}

First, Crosscheck checks if an activity failed to schedule due to an empty set of final valid intervals. 
While each activity $A_i$ is being scheduled, valid intervals are computed for the following constraints. 
\begin{itemize}
    \item Allowed start time windows/Execution ($T_i$)
    \item Dependencies ($D_i$)
    \item Unit Resources ($R_i$)
    \item State Requirements ($SR_i$)
    \item State Effects ($SE_i$)
    \item Data Volume ($dv_i$)
    \item UHF interactions ($U_i$)
\end{itemize}

The valid intervals for each of the above constraints are intersected together to find the final valid intervals for the activity. If the resulting set of valid intervals is empty, then the activity cannot be scheduled and further analysis must be conducted in order to find out which constraints were unable to be satisfied. Algorithm \ref{alg:failure-cause} describes the recursive algorithm Crosscheck uses in order to identify which constraints caused the activity to fail to schedule. 

\algdef{SE}[DOWHILE]{Do}{doWhile}{\algorithmicdo}[1]{\algorithmicwhile\ #1}%

{
\setlength\intextsep{2pt}
\begin{algorithm}[t]
\small
  \caption{Find Failure Cause from Constraint Intervals}
  \label{alg:failure-cause}
  \begin{algorithmic}[1]
  
  \Function{FindInvalidIntersections}{$VI_{prev}$, $VI_{list}$, $depth$, $dur$, $prevC$, $failedC$}
      \ForAll {$ind \in VI_{list}.size() - 1$} 
        \State $VI_{curr} \gets VI_{list}[ind]$
        \State $intersections \gets intersect(VI_{prev}, VI_{curr}, dur)$
        
         \If {$intersections.size() = 0 \land depth = 0$}
            \State $failedC.insert(prevC + \{curr\})$
         \ElsIf {$depth > 0$}
            \State \Call{FindInvalidIntersections}{$intersections$, $VI_{list}$[$ind+1$:], $depth-1$, $dur$, $prevC$ + $\{curr\}$, $failedC$}
         \EndIf
      \EndFor
  \EndFunction

   \Require
     \Statex $VI_{list}$: List of valid intervals for each constraint. $VI_c$ corresponds to the valid intervals for constraint $c$
     \Statex $dur$: Duration of activity
     \Statex $planBounds$: start time and end time of the plan
  \Ensure
     \Statex $failedC$: Minimal list of constraint types whose valid intervals intersected together result in activity failing to schedule.
     
  \State $depth \gets 0$
  \Do 
    \State $VI_{prev} \gets \{planBounds\}$ 
    \State $failedC \gets \{\}$
    \State $prevC \gets \{\}$ 
    \State \Call{FindInvalidIntersections}{$VI_{prev}$, $VI_{list}$, $depth$, 
    $dur$, $prevC$, $failedC$}
    \State {$depth \gets depth + 1$}
  \doWhile{$failedC \neq \{\}$}
    \State \Return $failedC$

  \end{algorithmic}
\end{algorithm}
}

First, Crosscheck checks if each individual constraint has valid intervals. If they all have at least one valid interval, then the algorithm will perform every possible pairwise intersection between the valid intervals. If each pair of constraints has valid intervals, then perform every three-way intersection, and so on. Ultimately through this process, Crosscheck will find the minimal set or sets of constraints that did not result in valid intervals after being intersected together, and thus, were the source of the activity failing to schedule.

In order to resolve the issue, one or more of the constraints that caused the activity to fail to schedule would need to be altered or a prior activity constraint would need to be altered in order to change the valid intervals for the activity that failed to schedule. 

In the worst case scenario, every subset of constraints has a non-empty intersection except the maximal set. To reach this point, each of the $2^n$ possible subsets would have to be checked for valid intersections. However, since this algorithm does not save computed intersections as depth increases, the complexity of the algorithm in the worst case (and making the assumption the intersection function is constant) is $O(n2^n)$ where $n$ is the number of constraints. Some constraint types may have multiple specific constraints (such as an activity having state requirements on multiple states). For each constraint type, the constraints are all intersected together before being input as a set of valid intervals to this intersection algorithm.
In practice with realistic plans, the maximum size of a set of constraints that do not have a valid intersection has been three. 

Once the first failure step and unsatisfiable constraints are determined, more specific information about which states or resources have constraints from both the failed activity and the activity scheduled prior to the first failure step are found and displayed to the user as the states/resources in conflict.

\subsection{Failure due to Plan-wide Constraints or Wake Sleep/Heater Scheduling}

An activity $A_i$ may fail to schedule to schedule for the following reasons during the second phase of scheduling the activity. Crosscheck adds detailed descriptions to inform the user of exactly which of the below constraints caused the activity prevented the activity from scheduling.

\begin{itemize}
    \item Insufficient energy to schedule the activity along with its required preheat ($P_i$), maintenance ($M_i$), and awake activities ($a_i$) without violating the minimum state of charge constraint, ($C\ _{soc}^{min}$). 
    \item Peak power of activity along with its required preheat ($P_i$), maintenance ($M_i$), and awake activities ($a_i$) would violate the maximum allowed peak power constraint ($C\ _{p\_power}^{max}$).
    \item Scheduling the required wake sleep activities would cause violation of the minimum required sleep time ($C\ _{sleep}^{min}$) or the minimum required awake time ($C\ _{awake}^{min}$).
    \item One or more of the required preheat activities $P_i$ would be scheduled outside of its operability window.
    \item One or more of the required preheat activities $P_i$ would be scheduled outside of the bounds of the plan. 
\end{itemize}

 At each time point considered for scheduling during this phase, one of the above reasons can be the cause of the failure, and this reason is output by the scheduler. As multiple time points may be considered for scheduling during this phase, there may be multiple reasons for the failure, and Crosscheck will indicate all of them. 
 
 In the following sections, we discuss how users will use Crosscheck to visualize the schedule, understand why activities failed to be scheduled, and which constraints need to be changed to achieve the desired schedule.

\section{Using Crosscheck}

Crosscheck enables the user to visualize the current state of the schedule at each scheduling step as well as information on why a specific activity failed to schedule. By knowing which constraints caused the activity to fail to schedule, users can understand how the constraints may be modified to enable scheduling of the activity. 

The schedule information at any given scheduling step is shown as a series of the following different types of timelines: 
\begin{itemize}
    \item \textit{Output schedule}. The activities in the output schedule have already been scheduling by the current scheduling step. That is, they are of higher priority (lower numerical value) than the activity being scheduled at the current step.
    \item \textit{Yet to be scheduled activities}. These activities are of lower priority than the activity being scheduled at the current step. Consequently they have not yet been scheduled. 
    \item \textit{Failed to schedule activities}.
    These activities have failed to schedule by the current scheduling step. 
    \item \textit{Energy profile}. The energy profile shows the total energy usage up until the current scheduling step.
    \item \textit{Peak Power profile}. The peak power profile shows the total peak power usage up until the current scheduling step.
\end{itemize}

In addition, users are able to view the following activity specific information by clicking on any activity in the timelines containing activities that successfully scheduled or failed to schedule: 
\begin{itemize}
    \item The partial schedule at the scheduling step during which the scheduler attempted to schedule the failed activity. 
    \item The valid intervals for each constraint as well as the final valid interval after intersecting the valid intervals for each constraint. 
\end{itemize}

For activities in the failed to schedule timeline, users can view detailed information on why the activity failed to schedule and which constraints need to be altered in order to allow the activity to schedule. 

\begin{figure*}[htb]
\centering
\includegraphics[width=\textwidth, height=20\baselineskip]{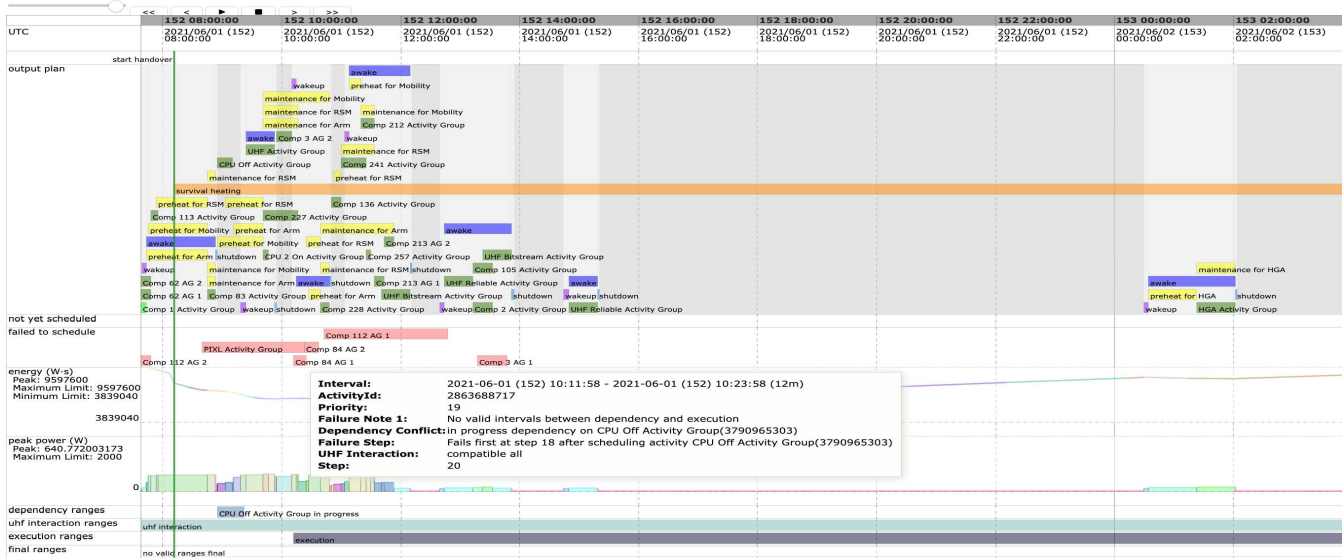}
\caption{Crosscheck allows users to view the schedule and specific information on why an activity failed to schedule as well as which constraints may need to be adjusted.}
\label{fig:vis}
\end{figure*}

Figure \ref{fig:vis} shows an example of a schedule shown in Crosscheck along with the various kinds of information described above.

In Figure \ref{fig:vis} an activity failed to schedule because there were no valid intervals after intersection the activity's valid intervals for its in progress dependency and execution constraints. This means that in order for the activity to schedule, on of two changes to the constraints must occur- 1) The activity's execution constraints need to be extended so that it can satisfy the dependency constraint as well or 2)the constraints of the activity that the failed activity was dependent on need to change so that it is scheduled during the execution intervals of the activity.

\subsection{Resource Usage}
Failures relating to depletable resources, such as energy, may not have one single activity causing the failure. Rather, it is the aggregate effects of multiple activities using the resource over separate times that would cause an activity to violate the minimum energy constraint. 

To aid in science planners remedying failures related to energy violations, they are able to view each activity's energy use up to the current time, for any time. If they were expecting an activity to be scheduled at a particular time, they can view all the energy users up to that time, in decreasing order of their energy use. This will allow them to potentially change the constraints on prior activities to change their final start times, allowing more energy to be available for the failed activity to use at that time. 

Similar information is available for peak power. As peak power is a non-depletable resource, it only matters which activities are using that resource at the current moment. Science planners can see which activities are using peak power at any time, as well as their peak power values, in decreasing order of the value. 

\begin{figure}[h]
\begin{center}
\subfloat[By clicking on the energy profile at a given time, users can view the energy consumed by each activity scheduled before that time. \label{fig:schedule_energy}]{%
\includegraphics[width=0.85\linewidth, height=8\baselineskip]{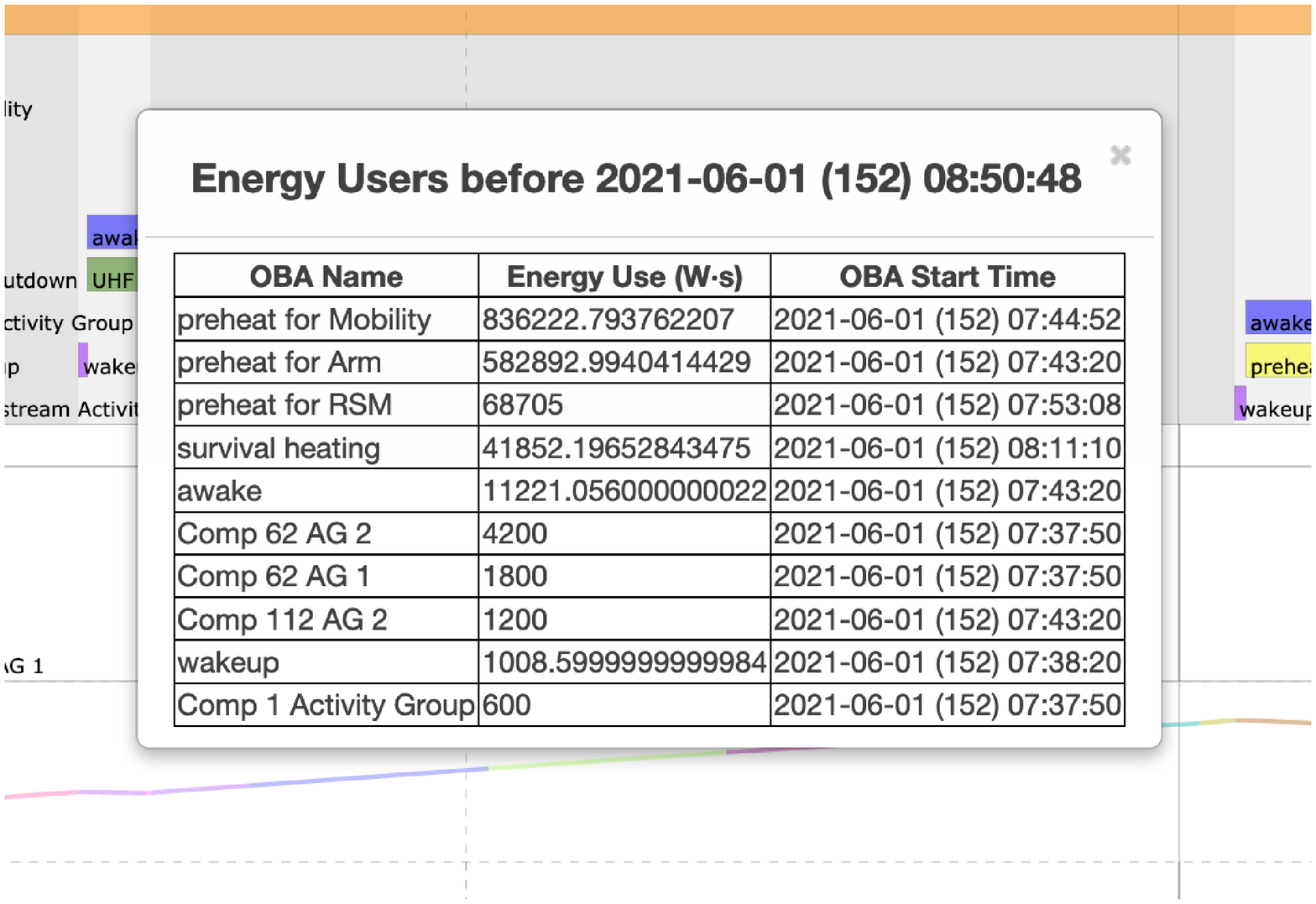}
}
\end{center}

\begin{center}
\subfloat[By clicking on the peak power profile at a given time, users can view the peak power for each activity at that time. \label{fig:schedule_peakpower}]{%
\includegraphics[width=0.85\linewidth, height=8\baselineskip]{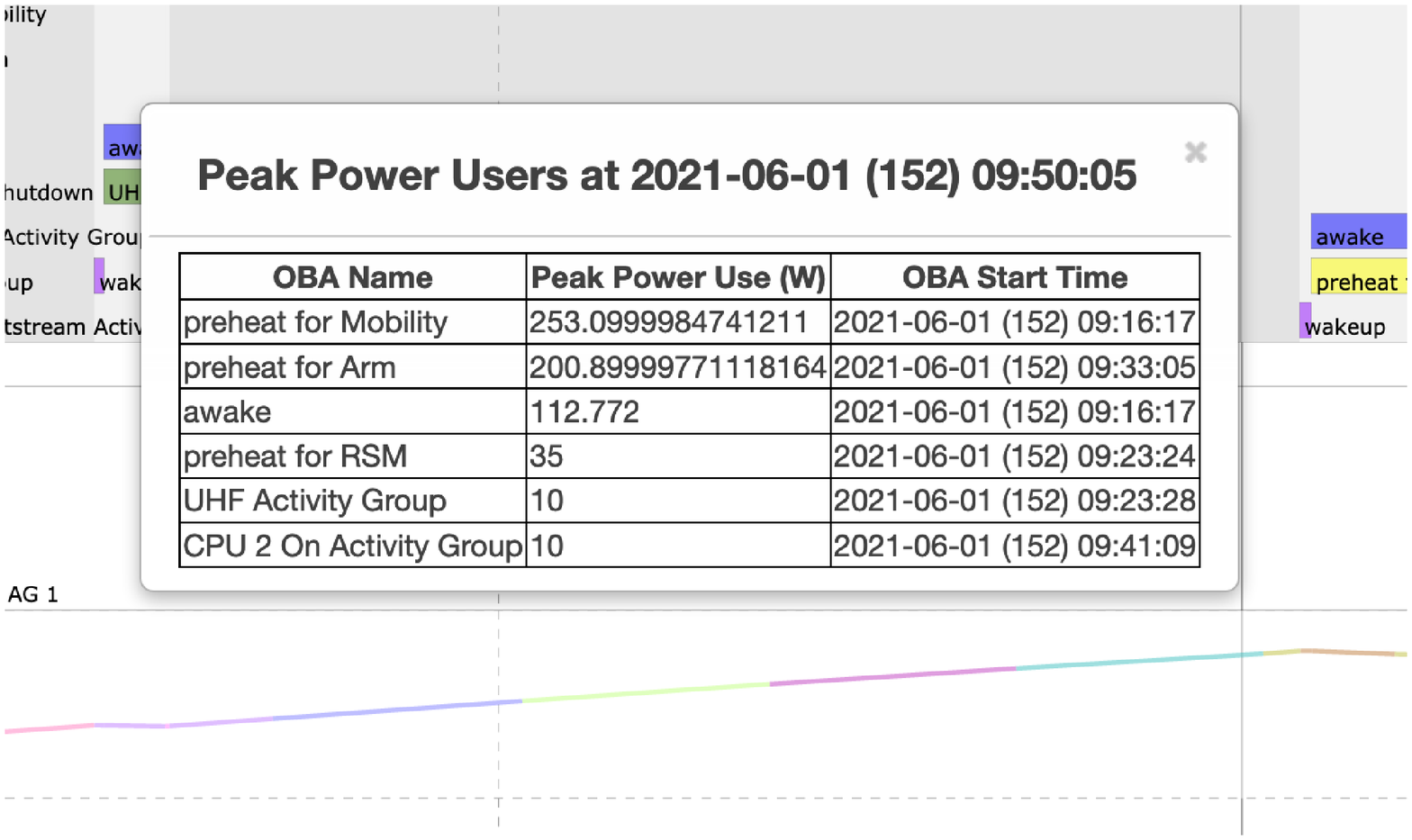}
}
\end{center}

\caption{Viewing energy and peak power usage in Crosscheck.}
\label{fig:resources}
\end{figure}

Figure \ref{fig:resources} shows how energy and peak power usage is visible in Crosscheck.

\squeezeup

\section{Crosscheck Examples of Identifying Scheduling Failures}

Crosscheck will provide reasons for why an activity failed to schedule as well as guidance on how constraints need to be changed in order to schedule the activity.

If an activity fails to schedule in the first phase of the scheduling algorithm, (the set of final valid intervals is empty), then Crosscheck will provide detailed information about which constraints must be altered in order to allow the activity to schedule. This information can be used by the users to modify the input activity constraints so that the scheduler generates the desired schedule.

\begin{figure}[h]
\begin{center}
\subfloat[The activity cannot schedule because there is no valid time where the activity can satisfy both the dependency and the uhf interaction constraints. \label{fig:valid}]{%
\includegraphics[width=0.95\linewidth, height=6\baselineskip]{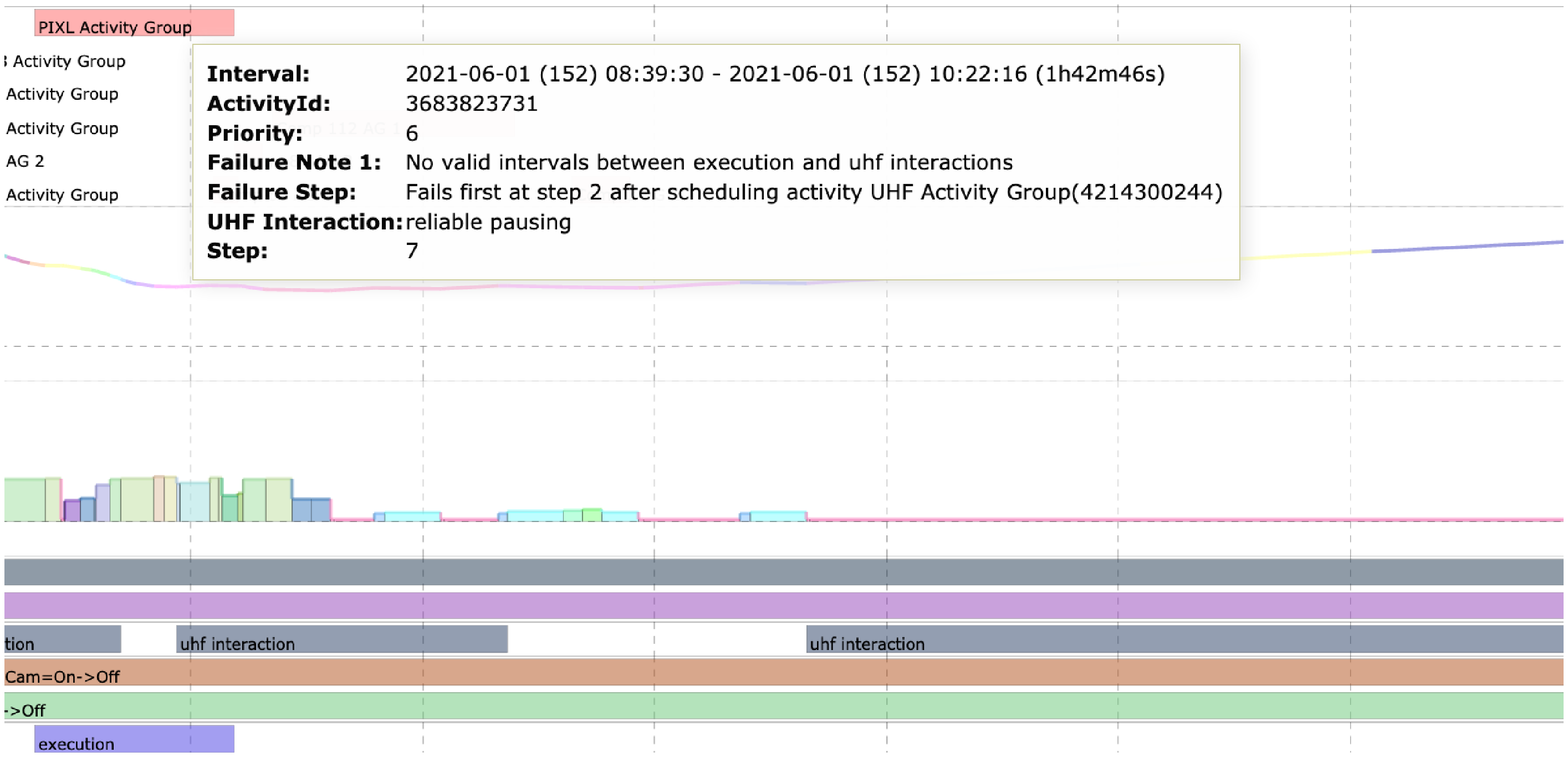}
}
\end{center}

\begin{center}
\subfloat[The activity cannot be scheduled because there is no valid intersection between the dependency valid intervals, unit resource valid intervals, and execution valid intervals. \label{fig:valid2}]{%
\includegraphics[width=0.95\linewidth, height=6\baselineskip]{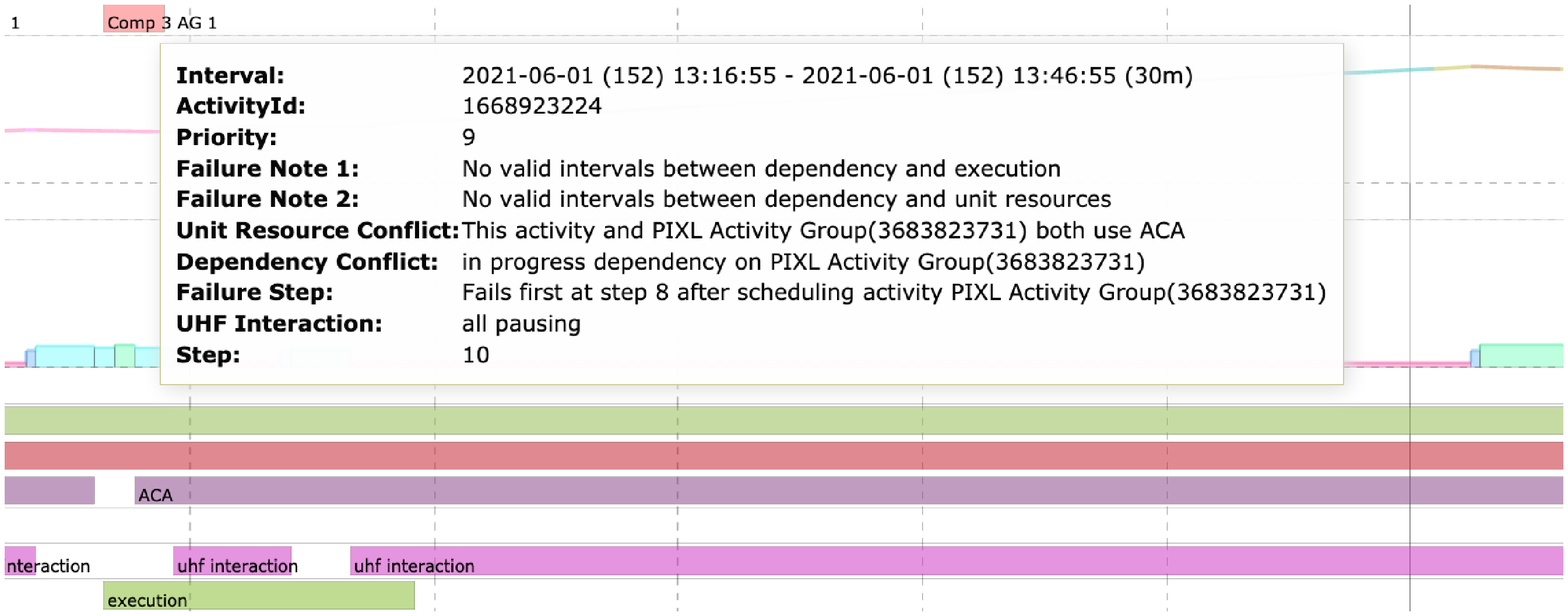}
}
\end{center}

\begin{center}
\subfloat[The activity cannot be scheduled because there is no valid time where the state requirement can be satisfied by a state effect of a prior activity. \label{fig:state}]{%
\includegraphics[width=0.95\linewidth, height=6\baselineskip]{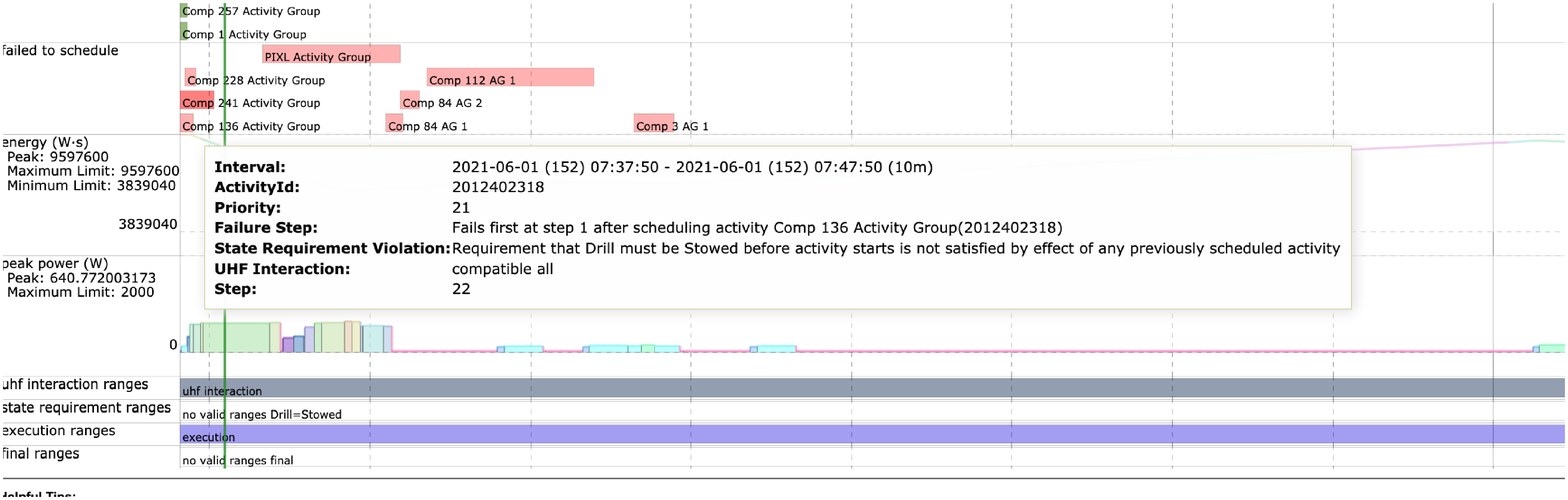}
}
\end{center}

\caption{Activities Fail to Schedule due to No Final Valid Intervals}
\label{fig:fail_phase_one}
\end{figure}

Figure \ref{fig:fail_phase_one} showcases several use cases of Crosscheck if an activity fails to schedule due to the activity having no final valid intervals. In Figure \ref{fig:valid}, as the Failure Note in Crosscheck indicates, the activity fails to schedule because there is no valid intersection between the activity's execution valid intervals and UHF interaction valid intervals. Consequently, in order for the activity to schedule, either its UHF interaction must change so that it does not have to execute in parallel with a UHF activity (which is less realistically plausible for science planners to be able to change) or the allowed start time windows of the activity must change. This is more feasible. If the allowed start time window of the activity is widened so that the execution valid intervals will have more overlap with the UHF interaction valid intervals, then the activity may be successfully scheduled. 

In Figure \ref{fig:valid2}, as indicated in Failure Notes 1 and 2, the activity fails to schedule because the activity's dependency valid intervals do not intersect with either the activity's execution valid intervals or its unit resource valid intervals. It is unrealistic for the activity's unit resources to be able to change, so either the dependency constraints or the execution constraints must be altered. While it is unlikely the science planner can remove the dependency from the activity that failed to schedule, it may be possible to alter the  constraints of the parent activity that it has a dependency on. If the parent activity then is scheduled elsewhere in the plan, then the valid intervals of the child activity would also be altered. It may also be possible to alter the activity's execution constraints so that the activity has allowed final valid intervals.

In Figure \ref{fig:state}, as indicated in the note for the State Requirement Violation, the activity failed to schedule because the requirement that the rover must have its drill stowed was not satisfied by any prior activity. This means that either another activity must be added that is scheduled temporally before this one so that the rover is in the required state or the incoming state of the rover must be the required state.

If the activity does have a nonempty set of final valid intervals, but fails to schedule during the second phase of the algorithm, then this means that the plan wide constraints (the minimum state of charge, the maximum peak power allowed, the minimum required sleep time, or the minimum required awake time) could not be satisfied while also accounting for the required heater and wake sleep activities. 

\begin{figure}[h]
\begin{center}
\subfloat[An activity may fail to schedule because scheduling it would cause the energy to go below the minimum allowed energy level \label{fig:energy}]{%
\includegraphics[width=0.95\linewidth, height=5\baselineskip]{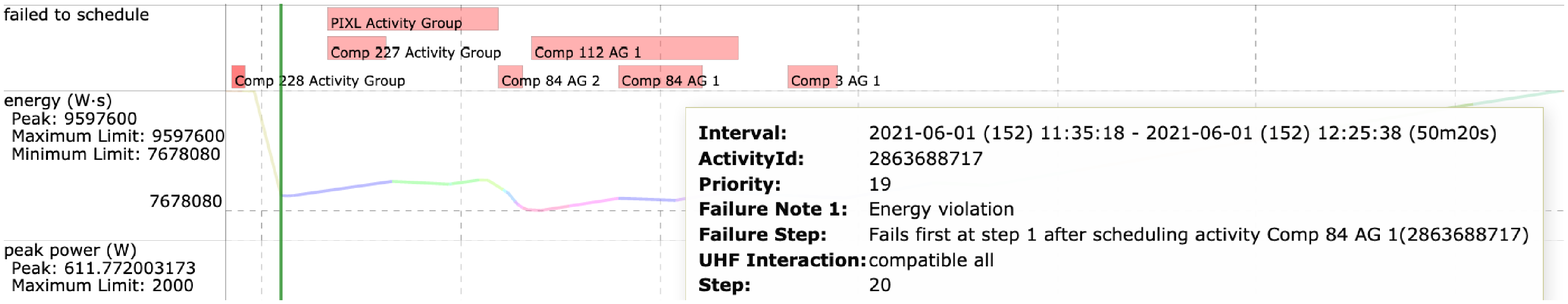}
}
\end{center}

\begin{center}
\subfloat[An activity may fail to schedule because the maximum allowed peak power would be violated\label{fig:peakpower}]{%
\includegraphics[width=0.895\linewidth, height=5\baselineskip]{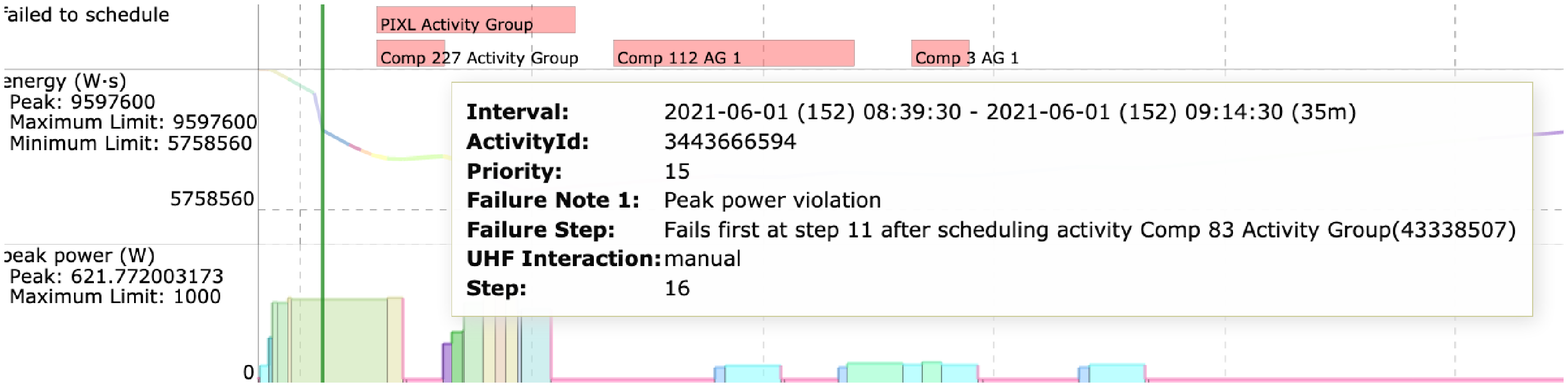}
}
\end{center}

\begin{center}
\subfloat[An activity may fail to schedule because the required preheat would be outside of its operability window\label{fig:operability}]{%
\includegraphics[width=0.95\linewidth, height=5\baselineskip]{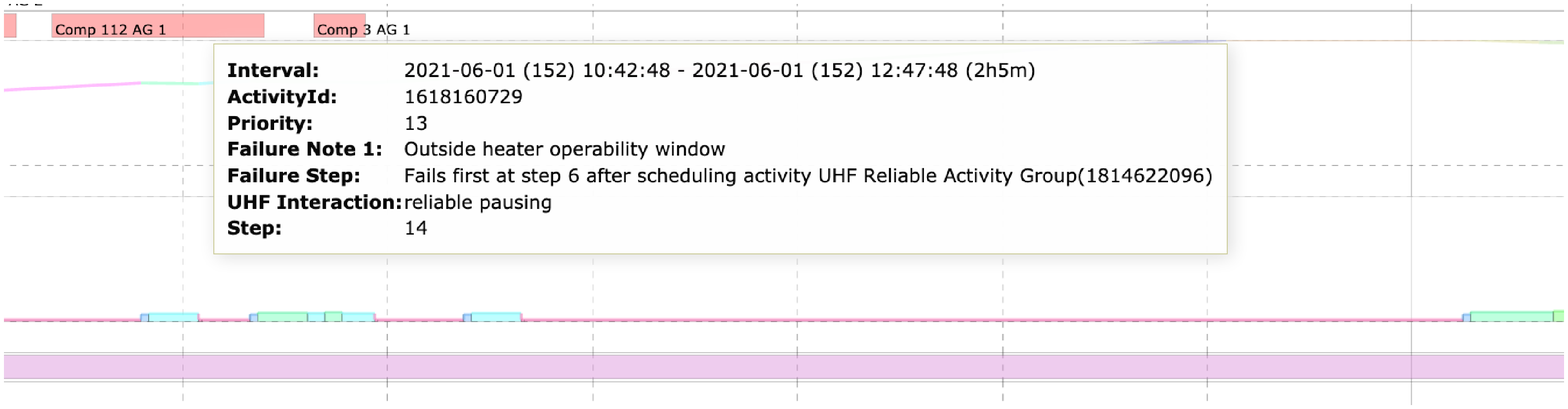}
}
\end{center}

\caption{Activities Fail to Schedule due to Plan Wide or Heater Specific Constraints}
\label{fig:fail_phase_two}
\end{figure}

In Figure \ref{fig:energy}, the activity fails to schedule because scheduling it along with its required wake sleep and heater activities would result an a violation of the minimum state of charge constraint. After knowing this information by reading Failure Note 1 generated by Crosscheck, the user may want to know which activities scheduled temporally before the one that failed were the most energy consumptive. As shown in Figure \ref{fig:schedule_energy}, the user can click on the energy timeline at a given time point and view the energy use of all activities occurring before that time point. For example, if the user clicked at a time on the energy timeline before the allowed execution valid interval of the activity, then they will be able to view the energy consumption of any activity that occurred before that time in order of most consumptive to least consumptive. Then, the user may be able to alter the constraints of one of the more consumptive activities so that it schedules later on in the plan when there is more energy available.

In Figure \ref{fig:peakpower}, the activity fails to schedule because scheduling it along with its required wake sleep and heater activities would result an a violation of the maximum allowed peak power constraint, as indicated by Failure Note 1. The user may want to know which activities were scheduled at the time the scheduler tried to schedule the activity and how much peak power those other activities used. As shown in \ref{fig:schedule_peakpower}, the user can click on the peak power timeline at a given time point and view the peak power use of all activities occurring at that time point. If the user clicks at a time on the energy timeline during the allowed execution valid intervals,  then they will be able to view the power consumption of any activity that occurs at that time in order of most consumptive to least consumptive. Then, the user may be able to alter the constraints of one of the more consumptive activities so that it can be scheduled later on in the sol when there is lower peak power usage.

\par
In Figure \ref{fig:operability}, the activity failed to schedule because the required preheat would be outside of its operability window. The science planner can clearly know this by viewing the failure note, and must alter the execution constraints of the activity so that the preheat would schedule within its operability window. 

\section{User Feedback}
Feedback from the intended users of Crosscheck (the science planners) has been positive. Throughout the development of the system, there have been weekly user testing sessions where plans are created, grounded using Copilot, and scheduling failures are diagnosed using Crosscheck. Users are in most cases able to clearly determine what needs to be changed in order for activities to successfully schedule through just looking at the notes given in the tooltips of the failed activity. Viewing the valid intervals provides them with more context on how the activity was scheduled, and allows them to see how execution constraints specifically may be changed to be compatible with the rest of the activity's constraints. 

Some of the terminology used in the Crosscheck visualizer is unfamiliar to those who do not have a scheduling background, such as the notion of valid intervals. Some background on how Copilot specifically works is needed to fully understand all the information being displayed, such as activities being scheduled in priority order. Further in-depth training will be required for new users for the system to be used when the rover lands.

\section{Related Work}
Bresina et al. \cite{bresina2005activity} describe an automated planning and scheduling system called MAPGEN which helps users create plans for the Mars Exploration Rovers. MAPGEN was a constraint-posting planner, and it explicitly indicated temporal flexibility, allowing users to drag activities until they reached their earliest or latest allowed start times. Unlike Crosscheck, it does not explicitly indicate why the planner could not achieve the desired result.

Both \cite{chakrabortiemerging} and \cite{fox2017explainable} conduct a survey of important prior work in the field of explaining planning. They discuss the  several important questions that arise when users interact with automated planning and scheduling systems as well as techniques that can help answer these questions. Crosscheck primarily focuses on answering why the scheduler did what it did given the activity constraints.

Ramaswamy et al. \cite{alper2019supporting} discuss a tool to visualize the numerous potential execution runs of a given input plan using the Mars 2020 ground scheduler. This tool provides information on how often an activity failed to execute over all execution runs. It also indicates how often activities executed in parallel over all execution runs and whether they switched order temporally. Crosscheck focuses on visualizing a single schedule and provides specific information on why an activity failed to schedule given the constraints provided.

The Rosetta Orbiter \cite{chien-rabideau-tran-et-al-IJCAI-2015} used automated scheduling to schedule science activities. 
A static display of the scheduling iterations is shown since ASPEN-RSSC also used a priority-based method to schedule activities, but it did not determine the higher priority iteration at which the activity would have been scheduled. The operators had to manually infer that information. Also the science campaigns only competed for the same resources; they did not set up preconditions such as dependencies for each other, meaning diagnosing failure to schedule was determining which competing campaign used the limiting resource(s).

Junker \cite{Junker2001QUICKXPLAINCD} describes an algorithm called QUICKXPLAIN for determining the minimal subset of constraints causing a conflict. QUICKXPLAIN has a better complexity than the algorithm presented in Failure Due To Constraint Valid Intervals, although this calculation is not a bottleneck in the overall runtime of Crosscheck.

\section{Future Work}
If the rover sleeps enough, the energy may reach the maximum allowed state of charge. If an activity fails to schedule due to violation of the minimum state of charge constraint after a time when the rover had maximum state of charge, any activities in the schedule prior to the time of maximum state of charge would not be at fault for the activity failing to schedule. When viewing resource users in Crosscheck, the energy users list contains all activities in the plan up to the time of interest, but not all of these activities are actually contributing to the energy value at that time. Future work includes showing only the relevant activities for an energy value at a given time. 
\par
Onboard the rover, activities may take more or less time than their predicted duration. If a certain set of execution durations results in activities failing to execute, then science planners may want to understand how they can change constraints so that there is a higher chance the activities will execute even if prior activities take longer than expected. When the mission begins using the onboard planner, Crosscheck can be used to analyze a given execution trace, determine why activities failed to execute for that set of execution durations, and help science planners understand which constraints need to be altered to ensure the activity will execute despite uncertainty in execution.

Although the calculation for determining the constraints that caused a conflict for an activity is not a bottleneck in Crosscheck, there are many avenues for improving the efficiency of this algorithm, including leveraging work from QUICKXPLAIN \cite{Junker2001QUICKXPLAINCD}. In the current implementation, there are repeated intersection calculations as the depth increases. Also, activities may not have all types of constraints, yet valid intervals are computed for them and input to the algorithm. In the future, only constraints that the activity actually has should be input. 

Future work also involves giving explicit suggestions for how constraints may be changed to allow an activity successfully schedule, in addition to the information already given in Crosscheck.  
\section{Conclusion}

In this paper we have described the automated scheduling system for the Mars 2020 Perseverance Rover mission and the Crosscheck explainable scheduling tool.  Crosscheck explains the scheduler behavior to users to build trust in the automated scheduling system. Crosscheck allows users to view the schedule at each scheduling iteration, gives detailed information on why an activity failed to schedule, and guides users on which constraints need to be altered in order for the activity to successfully schedule. This information will not only allow users to understand how to change activity constraints to achieve the desired schedule, but also helps users to build a model of scheduler behavior.

\bibliographystyle{aaai}
\bibliography{references}

\end{document}